\documentclass{article}
\usepackage[utf8]{inputenc}
\usepackage{rotating}
\usepackage{float}
\usepackage{wrapfig}
\usepackage{units}
\usepackage{amsmath}
\usepackage{graphicx}
\usepackage[authoryear]{natbib}
\PassOptionsToPackage{normalem}{ulem}
\usepackage{ulem}

\makeatletter

\providecommand{\tabularnewline}{\\}
\newcommand{\lyxdot}{.}

\usepackage{iclr2016_conference, times}
\usepackage{hyperref}
\usepackage{url}
\usepackage{xspace}

\title{A convnet for non-maximum suppression}

\author{Jan Hosang, Rodrigo Benenson, and Bernt Schiele
\\
Max-Planck Institute for Informatics\\
Saarbr\"ucken, Germany\\
\texttt{\{jhosang, benenson, schiele\}@mpi-inf.mpg.de}
}

\@ifundefined{showcaptionsetup}{}{%
 \PassOptionsToPackage{caption=false}{subfig}}
\usepackage{subfig}
\makeatother

\begin{document}
\maketitle

\newcommand{\NMS}{NMS\xspace}
\begin{abstract}
Non-maximum suppression (\NMS) is used in virtually all state-of-the-art
object detection pipelines. While essential object detection ingredients
such as features, classifiers, and proposal methods have been extensively
researched surprisingly little work has aimed to systematically address
\NMS. The de-facto standard for \NMS is based on greedy clustering
with a fixed distance threshold, which forces to trade-off recall
versus precision. We propose a convnet designed to perform \NMS of
a given set of detections. We report experiments on a synthetic setup,
and results on crowded pedestrian detection scenes. Our approach overcomes
the intrinsic limitations of greedy \NMS, obtaining better recall
and precision.
\end{abstract}
\makeatletter 
\renewcommand{\paragraph}{%
\@startsection{paragraph}{4}%
{\z@}{0.5ex \@plus 1ex \@minus .2ex}{-0.5em}%
{\normalfont \normalsize \bfseries}%
}
\makeatother

\section{\label{sec:Introduction}Introduction}

The bulk of current object detection pipelines are based on three
steps: 1) propose a set of windows (either via sliding window, or
object proposals), 2) score each window via a properly trained classifier,
3) remove overlapping detections (non-maximum suppression). The popular
DPM \citep{Felzenszwalb2010Pami} and R-CNN family \citep{Girshick2014Cvpr,Girshick2015IccvFastRCNN,Ren2015NipsFasterRCNN}
follow this approach. Both object proposals \citep{Hosang2015Pami}
and detection classifiers \citep{Russakovsky2015IjcvImageNet} have
received enormous attention, while non-maximum suppression (\NMS)
has been seldom addressed. The de-facto standard for \NMS consists
of greedily merging the higher scoring windows with lower scoring
windows if they overlap enough (e.g. intersection-over-union $\mbox{IoU}>0.5$),
which we call GreedyNMS in the following.

\begin{figure}
\begin{centering}
\includegraphics[width=0.9\textwidth]{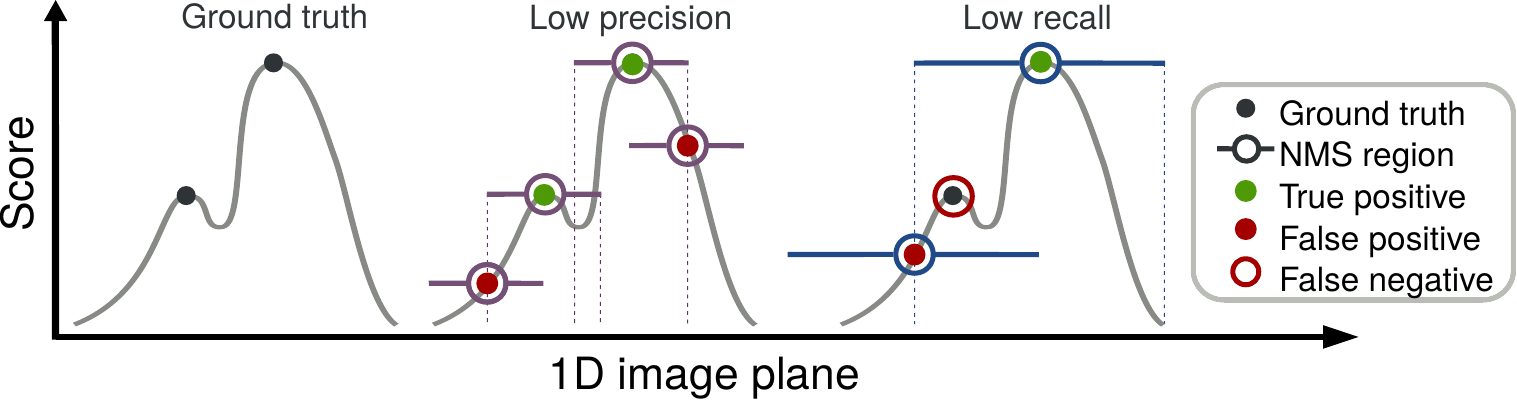}
\par\end{centering}

\caption{\label{fig:greedy-nms-issues}1D illustration of the GreedyNMS shortcomings.
Black dots indicate true objects, grey curve is the detector response,
green dots are true positives, red dots/circles are false positives/negatives.
Whenever the merge criterion is too narrow false positives will be
introduced (low precision, middle case), when the merge criterion
is too wide there are missed detections (low recall, right case).
A fixed merging criterion is doomed to fail across different situations.}
\end{figure}

GreedyNMS is popular because it is conceptually simple, fast, and
for most tasks results in satisfactory detection quality. Despite
its popularity, GreedyNMS has important shortcomings. As illustrated
in figure \ref{fig:greedy-nms-issues} (and also shown experimentally
in section \ref{sec:pets-experiments}) GreedyNMS trades off precision
vs.~recall. If the $\mbox{IoU}$ threshold is too large (too strict)
then not enough surrounding detections are suppressed, high scoring
false positives are introduced and precision suffers; if the $\mbox{IoU}$
threshold is too low (too loose) then multiple true positives are
merged together and the recall suffers (right-most case in figure
\ref{fig:greedy-nms-issues}). For any $\mbox{IoU}$ threshold, GreedyNMS
is sacrificing precision or recall. One can do better than this by
leveraging the full signal of the score map (statistics of the surrounding
detections) rather than blindly applying a fixed policy everywhere
in the image.

Current object detectors are becoming surprisingly effective on both
general (e.g. Pascal VOC, MS COCO) and specific objects detection
(e.g pedestrians, faces). The oracle analyses for ``perfect \NMS''
from \citet[table 5]{Hosang2015Pami} and \citet[figure 12]{Parikh2011Nipsw}
both indicate that \NMS accounts for almost a quarter of the remaining
detection problem.

GreedyNMS is, strictly speaking, not a ``non-maximum suppression''
method since it focuses on removing overlapping detections, ignoring
if a detection is a local maximum or not. To improve detections, we
would like to prune false positives (and keep true positives). Instead
of doing hard pruning decisions, we design our network to make soft
decisions by re-scoring (re-ranking) the input detection windows.
Our re-scoring is final, and no post-processing is done afterwards,
thus the resulting score maps must be very ``peaky''. We call our
proposed network ``Tyrolean network''\footnote{Tyrolean because ``it likes to see peaky score maps''.},
abbreviated Tnet.

\paragraph{Contribution}

We are the first to show that a convnet can be trained and used to
overcome the limitations of GreedyNMS. Our experiments demonstrate
that, across different occlusion levels, the Tyrolean network (Tnet)
performs strictly better than GreedyNMS at \emph{any} IoU threshold.

As an interesting scenario for \NMS, we report results over crowded
pedestrian scenes. Our Tnet is focused on the \NMS task, it does
not directly access image content (thus does not act as ``second
detector''), does not use external training data, and provides better
results than auto-context \citep{Tu2010Pami}.

\subsection{\label{sub:Related-work}Related work}

Despite being a core ingredient of detection, non-maximum suppression
has received little attention compared to feature design, classifier
design, and object proposals.

\paragraph{Clustering detections}

The decade old greedy \NMS (GreedyNMS) is used in popular detectors
such as V\&J \citep{Viola2004Ijvc}, DPM \citep{Felzenszwalb2010Pami},
and is still used in the state-of-the-art R-CNN detector family \citep{Girshick2014Cvpr,Girshick2015IccvFastRCNN,Ren2015NipsFasterRCNN}.
Alternatives such as mean-shift clustering \citep{Dalal2005Cvpr,Wojek2008Pr},
agglomerative clustering \citep{Bourdev2010Eccv}, and heuristic variants
\citep{Sermanet2014Iclr} have been considered, but they have yet
to show consistent gains. Recently \citet{Tang2015Cvpr,Rothe2015Accv}
proposed principled clustering approaches that provide globally optimal
solutions, however the results reached are on par, but do not surpass,
GreedyNMS.

\paragraph{Linking detections to pixels}

The Hough voting framework enables more sophisticated reasoning amongst
conflicting detections by linking the detections to local image evidence
\citep{Leibe2008Ijcv,Barinova2012Pami,Wohlhart2012Accv}. Hough voting
itself, however, provides low detection accuracy. \citet{Yao2012Cvpr}
and \citet{Dai2015Cvpr} refine detections by linking them with semantic
labelling; while \citet{Yan2015Cvpr} side-steps \NMS all-together
by defining the detection task directly as a labelling problem. These
approaches arguably propose a sound formulation of the detection problem,
however they rely on semantic labelling/image segmentation. Our system
can operate directly on bounding box detections.

\paragraph{Co-occurrence}

To better handle dense crowds or common object pairs, it has been
proposed to use specialized 2-object detectors \citep{Sadeghi2011Cvpr,Tang2012Bmvc,Ouyang2013Cvpr},
which then require a more careful \NMS strategy to merge single-object
with double-object detections. Similarly \citet{Rodriguez2011Iccv}
proposed to adapt the \NMS threshold using crowd density estimation.
Our approach is directly learning based (no hand-crafted 2-objects
or density estimators), and does not use additional image information.\\
\citet{Desai2011Ijcv} considered improving the \NMS procedure by
considering the spatial arrangement between detections of different
classes. The feature for spatial arrangement are fully hand-crafted,
while we let a convnet learn the spatial arrangement patterns. We
focus here on the single class case (albeit our approach could be
extended to handle multiple classes).

\paragraph{Speed}

\citet{Neubeck2006Icpr} discuss how to improve \NMS speed, without
aiming to improve quality. Here we aim at improving quality, while
still having practical speeds.

\paragraph{Auto-context}

uses the local \citep{Tu2010Pami,Chen2013CvprMoco} or global information
\citep{Vezhnevets2015Bmvc} on the image to re-score detections. Albeit
such approaches do improve detection quality, they still require a
final \NMS processing step. Our convnet does re-score detections,
but at the same time outputs a score map that does not require further
processing. We provide experiments (section \ref{sec:pets-experiments})
that show improved performance over auto-context.

\paragraph{Convnets and \NMS}

Recently a few works have linked convnets and \NMS. Detection convnets
are commonly trained unaware of the \NMS post-processing step; \citet{Wan2015Cvpr}
connected \NMS with the loss to train the detection convnet, making
the training truly end-to-end. The used \NMS is greedy and with fixed
parameters. \citet{Stewart2015Arxiv} propose to use an LSTM to decide
how many detections should be considered in a local region. The detections
amongst the regions are then merged via traditional \NMS. In contrast
our convnet runs in a fully convolutional mode and requires no post-processing.
Both \citet{Wan2015Cvpr} and \citet{Stewart2015Arxiv} served as
inspiration for the design of our training loss. To the best of our
knowledge our Tnet is the first network explicitly designed to replace
the final \NMS stage.\\
This work focuses on NMS itself, our proposed network is independent
of the detections source (it can be from a convnet or not). We report
experiment applied over DPM \citep{Felzenszwalb2010Pami} detections.

\section{\label{sec:Base-convnet}Base Tyrolean network}

\begin{figure}
\begin{centering}
\includegraphics[width=1\textwidth]{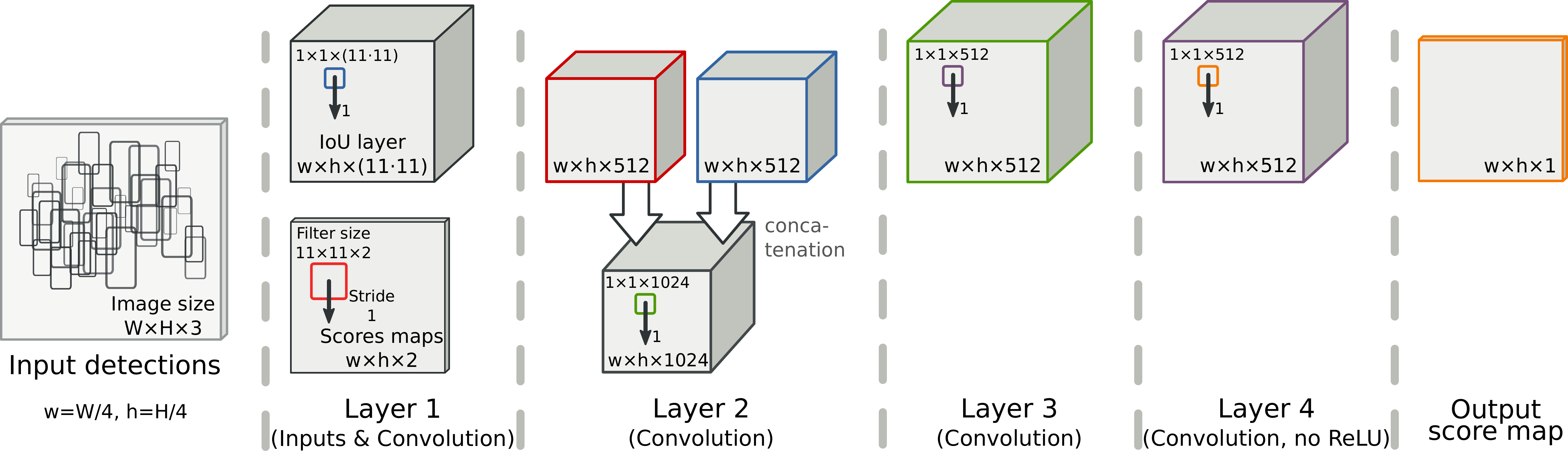}
\par\end{centering}

\caption{\label{fig:Tnet}Base architecture of our Tyrolean network (Tnet).
See text for the variants considered. Each grey box is a feature map,
its dimensions are indicated at its bottom, the coloured square indicates
the convolutional filters size, the stride is marked next to the downward
arrow. Unless indicated each layer uses ReLu non-linearities. No max-pooling
nor local normalization is used. Any other processing marked in grey
text.}
\end{figure}

The main intuition behind our proposed Tyrolean network (Tnet) is
that the score map of a detector together with a map that represents
the overlap between neighbouring hypotheses contains valuable information
to perform better \NMS than GreedyNMS (also see figure \ref{fig:greedy-nms-issues}).
Thus, our network is a traditional convnet but with access to two
slightly unusual inputs (described below), namely score map information
and $\mbox{IoU}$ maps. Figure \ref{fig:Tnet} shows the overall network.
The parameter range (number of layers, number of units per layer)
is inspired by AlexNet \citep{Krizhevsky2012Nips} and VGG \citep{Simonyan2015Iclr}.
In our base Tnet the first stage applies $512$ $11\times11$ filters
over each input layer, and $512$ $1\times1$ filters are applied
on layers 2, 3, and 4. ReLU non-linearities are used after each layer
but the last one.

The network is trained and tested in a fully convolutional fashion.
It uses the same information as GreedyNMS, and does not access the
image pixels directly. The required training data are only a set
of object detections (before NMS), and the ground truth bounding boxes
of the dataset.

\paragraph{Input grid }

As preprocessing all detections in an image are mapped into a 2d grid
(based on their centre location). If more than one detection falls
into the same cell, we keep only the highest scoring detection. Each
cell in the grid is associated with a detection bounding box and score.
We use cells of $4\times4\ \mbox{pixels}$, thus an input image of
size $\mbox{W}\times\mbox{H}$ will be mapped to input layers of size
$\mbox{w}\times\mbox{h}=\nicefrac{\mbox{W}}{4}\times\nicefrac{\mbox{\mbox{H}}}{4}$.
Since the cells are small, mapping detections to the input grid has
minimal impact on the \NMS procedure. In preliminary experiments
we validated that: a) we can at least recover the performance of GreedyNMS
(applying GreedyNMS over the input grid provides the same results
as directly applying GreedyNMS), b) the detection recall stays the
same (after mapping to the input grid the overall recall is essentially
identical to the raw detections).\\
The current incarnation of Tnet can handle mild changes in scale amongst
neighbouring detections. We report experiments with detections over
a $3\times$ scale range (most boxes have $[50,\,150]\ \mbox{pixel}$
height). The overall approach can handle a wider range of scale differences
with simple modifications (making the grid 3d, i.e. x-y-scale), which
is left for future work.

\paragraph{$\mbox{IoU}$ layer}

In order to reason about neighbouring detection boxes (or segments)
we feed Tnet with $\mbox{IoU}$ values. For each location we consider
a $11\times11=121$ neighbourhood, thus the input $\mbox{IoU}$ layer
has $\mbox{w}\times\mbox{h}\times121$ values. Together the cell size
and neighbourhood size should provide the Tnet with sufficient information
about surroundings of a detection, where this choice depends on the
object sizes in the image and the expected object density and thus
are application dependent.

\paragraph{Score maps layer}

To reason about the detection confidence, we feed Tnet with the raw
detection score map (once mapped to the input grid). The \NMS task
involves ranking operations which are not easily computed by linear
and relu ($\max(\cdot,\,0)$) operators. To ease the task we also
feed the Tnet with score maps resulting from GreedyNMS at multiple
$\mbox{IoU}$ thresholds. All score maps are stacked as a multi-channel
input image and feed into the network. $\mbox{S}\negthinspace\left(\tau\right)$
denotes a score map resulting from applying GreedyNMS with $\mbox{IoU\ensuremath{\geq\tau}}$,
$\mbox{S}\negthinspace\left(\tau_{1},\,\tau_{2}\right)$ denotes a
two channels map ($\mbox{S}\negthinspace\left(\tau_{1}\right)$ and
$\mbox{S}\negthinspace\left(\tau_{2}\right)$ stacked). Note that
$\mbox{S}\negthinspace\left(1\right)$ returns the raw detection score
map. Our base Tnet uses $\mbox{S}\negthinspace\left(1,\,0.3\right)$
which has dimensionality $\mbox{w}\times\mbox{h}\times2$ (see figure
\ref{fig:Tnet}). The convolutional filters applied over the score
maps input have the same size as the $\mbox{IoU}$ layer neighbourhood
($11\times11\ \mbox{cells}$).

Tnet is then responsible for interpreting the multiple score maps
and the $\mbox{IoU}$ layer, and make the best local decision. Our
Tnet operates in a fully feed-forward convolutional manner. Each location
is visited only once, and the decision is final. In other words, for
each location the Tnet has to decide if a particular detection score
corresponds to a correct detection or will be suppressed by a neighbouring
detection in a single feed-forward path.

\paragraph{Parameter rules of thumb}

Figure \ref{fig:Tnet} indicates the base parameters used. Preliminary
experiments indicated that removing some of the top layers has a
clear negative impact on the network performance, while the width
of these layers is not that important ($512$, $1\,024$, $2\,048$
filters in layers 2,3, and 4 shows a slow increase in quality). Having
a high enough resolution in the input grid is critical, while keeping
a small enough number of convolutions over the inputs allows to keep
the number of model parameters under control. During training data
augmentation is necessary to avoid overfitting. The training procedure
is discussed in more detail in \ref{sub:Training}, while experimental
results for some parameters variants are reported in section \ref{sec:pets-experiments}.

\paragraph{Input variants}

In the experiments in the next sections we consider different input
variants. The $\mbox{IoU}$ layer values can be computed over bounding
boxes (regressed by the sliding window detector), or over estimated
instance segments \citep{Pinheiro2015ArxivDeepMask}.\\
Similarly, for the score maps we consider different numbers of GreedyNMS
thresholds, which changes the dimensionality of the input score map
layer.\\
In all cases we expect the Tnet to improve over a fixed threshold
GreedyNMS by discovering patterns in the detector score maps and $\mbox{IoU}$
arrangements that enable to do adaptive \NMS decisions.

\subsection{\label{sub:Training}Training procedure}

Typically detectors are trained as classifiers on a set of positive
and negative windows, determined by the IoU between detection and
object annotation. When doing so the spatial relation between detector
outputs and the annotations is being neglected and little work has
addressed this.The DPM \citep{Felzenszwalb2010Pami,Pepik2015Pami}
includes structured output learning to ensure the detection score
falls off linearly with the overlap between detector window and annotation.
\citet{Wan2015Cvpr} explicitly include a fixed \NMS procedure into
the network loss during training so the detector is tuned towards
the \NMS at test time. We adopt from \citet{Stewart2015Arxiv} the
idea of computing the loss by matching detections to annotations,
but instead of regressing bounding boxes at every location we predict
new detection scores that are high for matched detections and low
everywhere else. In contrast to the conventional wisdom of training
the detector to have a smooth score decrease around positive instances,
we declare a detection right next to a true positive to be a negative
training sample as long as it is not matched to an annotation. We
do this because our network must itself perform \NMS.

\paragraph{Training loss}

Our goal is to reduce the score of all detections that belong to the
same person, except exactly one of them. To that end, we match every
annotation to all detections that overlap at least $0.5$ IoU and
choose the maximum scoring detection among them as the one positive
training sample. All other detections are negative training examples.
This yields a label $y_{p}$ for every location $p$ in the input
grid (see previous section). Since background detections are much
more frequent than true positives, it is necessary to weight the loss
terms to balance the two. We use the weighted logistic loss

\begin{equation}
L(\mathbf{x})=\frac{1}{\sum_{p\in G}w_{y_{p}}}\sum_{p\in G}w_{y_{p}}\log\left(1+e^{-y_{p}f(x_{p})}\right)\label{eq:loss}
\end{equation}
where $x_{p}$ is the feature descriptor at position $p$, $f(x_{p})$
is the output of the network at position $p$. The weights $w_{y_{p}}$
are chosen so that both classes have the same weight either per frame
or globally on the entire dataset (denoted by $w_{f}$ and $w_{g}$
respectively). Since we have a one-to-one correspondence between input
grid cells and labels it is straight forward to train a fully convolutional
network to minimize this loss.

\paragraph{Relaxed loss}

It is impossible for the network to recover from certain mistakes
that are already present in the input detections. For example, false
positives on the background might be impossible to tell apart from
true positives since the network does not have access to the image
and only sees detection scores and overlaps between detections. On
the other hand detections of distinct objects with high overlap can
be hard to detect since the detections can assign low scores to barely
visible objects. It proved beneficial to assign lower weight to these
cases, which we call the \emph{relaxed loss}. We declare negative
samples to be hard if the corresponding detections are not suppressed
by a 0.3 \NMS and true positives to be hard if they are suppressed
by a 0.3 \NMS on the annotations with the matched detection scores.
The weight of hard examples is decreased by a factor of $r$. Our
base Tnet uses $r=1$ (non-relaxed) with weighting strategy $w_{f}$,
and section \ref{sec:pets-experiments} reports results for other
$r$ values and $w_{g}$.

\paragraph{Training parameters}

The model is trained from scratch, randomly initialized with MSRA
\citep{He2015iccv}, and optimized with Adam \citep{Kingma2015iclr}.
We use a learning rate of $10^{-4}$, a weight decay of $5\cdot10^{-5}$,
a momentum of 0.9, and gradient clipping at $1\,000$. The model is
trained for $100\,000$ iterations with one image per iteration. All
experiments are implemented with the Caffe framework \citep{Jia2014caffe}.

As pointed out in \citet{Mathias2014Eccv} the threshold for GreedyNMS
requires to be carefully selected on the validation set of each task,
the commonly used default $\mbox{IoU}>0.5$ can severely underperform.
Other \NMS approaches such as \citep{Tang2015Cvpr,Rothe2015Accv}
also require training data to be adjusted. When maximizing performance
in cluttered scenes is important, training a Tnet is thus not a particularly
heavy burden. Currently, training our base Tnet on un-optimized CPU
and GPU code takes a day.

\section{\label{sec:mnist-experiments}Controlled setup experiments}

\NMS is usually the last stage of an entire detection pipeline. Therefore,
in an initial set of experiments, we want to understand the problem
independent of a specific detector and abstract away the particularities
of a given dataset.

\subsection{\label{sub:oMNIST-dataset}oMNIST dataset}

If all objects appeared alone in the images, \NMS would be trivial.
The core issue for \NMS is deciding if two local maxima in the detection
score map correspond to only one object or to multiple ones. To investigate
this core aspect we create the oMNIST (``overlapping MNIST'') toy
dataset. This data does not aim at being particularly realistic, but
rather to enable a detailed analysis of the \NMS problem.
\begin{figure}
\begin{centering}
\begin{tabular}{lc}
Image & \includegraphics[width=0.8\textwidth]{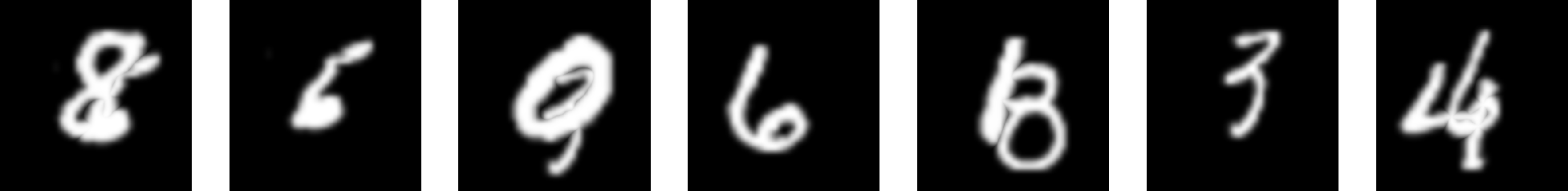}\tabularnewline
\vspace{-0.7em}
 & \tabularnewline
Score map & \includegraphics[width=0.8\textwidth]{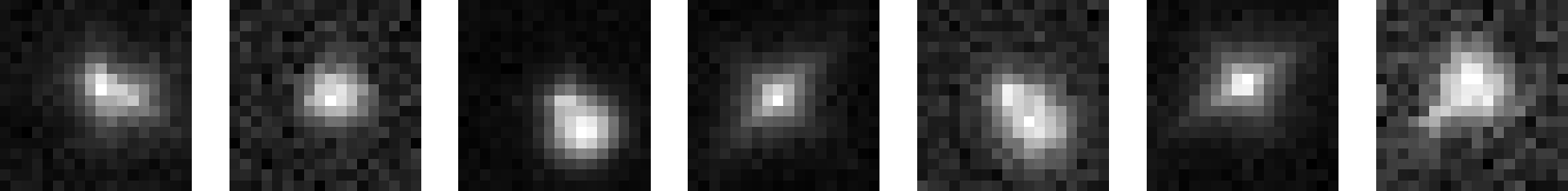}\tabularnewline
\end{tabular}
\par\end{centering}

\caption{\label{fig:oMnist}Example data from our controlled experiments setup.
The convnet must decide if one or two digits are present (and predict
is their exact location) while using only a local view of score and
$\mbox{IoU}$ maps (no access to the input image). }
\end{figure}

Each image is composed of one or two MNIST digits. To emphasise the
occlusion cases, we sample $\nicefrac{1}{5}$ single digits, and $\nicefrac{4}{5}$
double digit cases. The digits are off-centre and when two digits
are present they overlap with bounding box $\mbox{IoU}\in[0.2,\,0.6]$.
We also mimic a detector by generating synthetic score maps. Each
ground truth digit location generates a perturbed bump with random
magnitude in the range $[1,\,9]$, random x-y scaling, rotation, a
small translation, and additive Gaussian noise. Albeit noisy, the
detector is ``ideal'' since its detection score remains high despite
strong occlusions. Figure \ref{fig:oMnist} shows examples of the
generated score maps and corresponding images. By design GreedyNMS
will have difficulties handling such cases (at any $\mbox{IoU}$ threshold). 

Other than score maps our convnet uses $\mbox{IoU}$ information between
neighbouring detections (like GreedyNMS). In our experiments we consider
using the perfect segmentation masks for $\mbox{IoU}$ (ideal case),
noisy segmentation masks, and the sliding window bounding boxes.\begin{wrapfigure}{O}{0.45\columnwidth}%
\vspace{-1em}

\begin{centering}
\begin{tabular}{cc}
\begin{turn}{90}
\hspace{0.6em}{\footnotesize{}Image}
\end{turn} & \hspace{-1em}\includegraphics[width=0.4\textwidth]{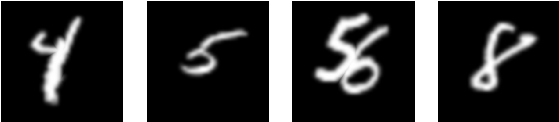}\tabularnewline
{\footnotesize{}\vspace{-1.2em}
} & \tabularnewline
\begin{turn}{90}
\hspace{-0em}{\footnotesize{}Score map}
\end{turn} & \hspace{-1em}\includegraphics[width=0.4\textwidth]{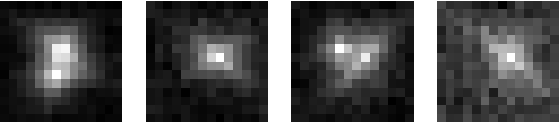}\tabularnewline
{\footnotesize{}\vspace{-1.1em}
} & \tabularnewline
\begin{turn}{90}
\hspace{-0.9em}{\footnotesize{}GreedyNMS}
\end{turn} & \hspace{-1em}\includegraphics[width=0.4\textwidth]{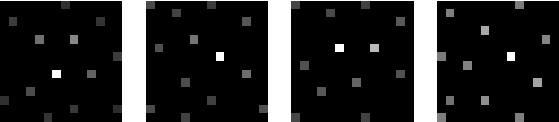}\tabularnewline
{\footnotesize{}\vspace{-0.8em}
} & \tabularnewline
\begin{turn}{90}
\hspace{1em}{\footnotesize{}Tnet}
\end{turn} & \hspace{-1em}\includegraphics[width=0.4\textwidth]{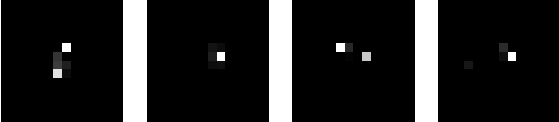}\tabularnewline
\end{tabular}\vspace{-0.5em}

\par\end{centering}

\caption{\label{fig:omnist-nms-tnet}Example score maps on the oMNIST dataset.
First to last row: Image, input score map, GreedyNMS $\mbox{IoU}>0.3$,
and Tnet $\mbox{IoU}\,\&\,\mbox{S}\negthinspace\left(1,\,0\rightarrow0.6\right)$.
GreedyNMS produces false positives and prunes true positives, while
Tnet correctly localize even very close digits.}

\vspace{-6em}
\end{wrapfigure}%

We generate a training/test split of $100\mbox{k}$/$10\mbox{k}$
images, kept fix amongst different experiments.

\subsection{\label{sub:mnist-results}Results}

\begin{figure*}[t]

\hspace*{\fill}%
\begin{minipage}[t]{0.63\columnwidth}%
\begin{figure}[H]
\begin{centering}
\includegraphics[width=1\textwidth]{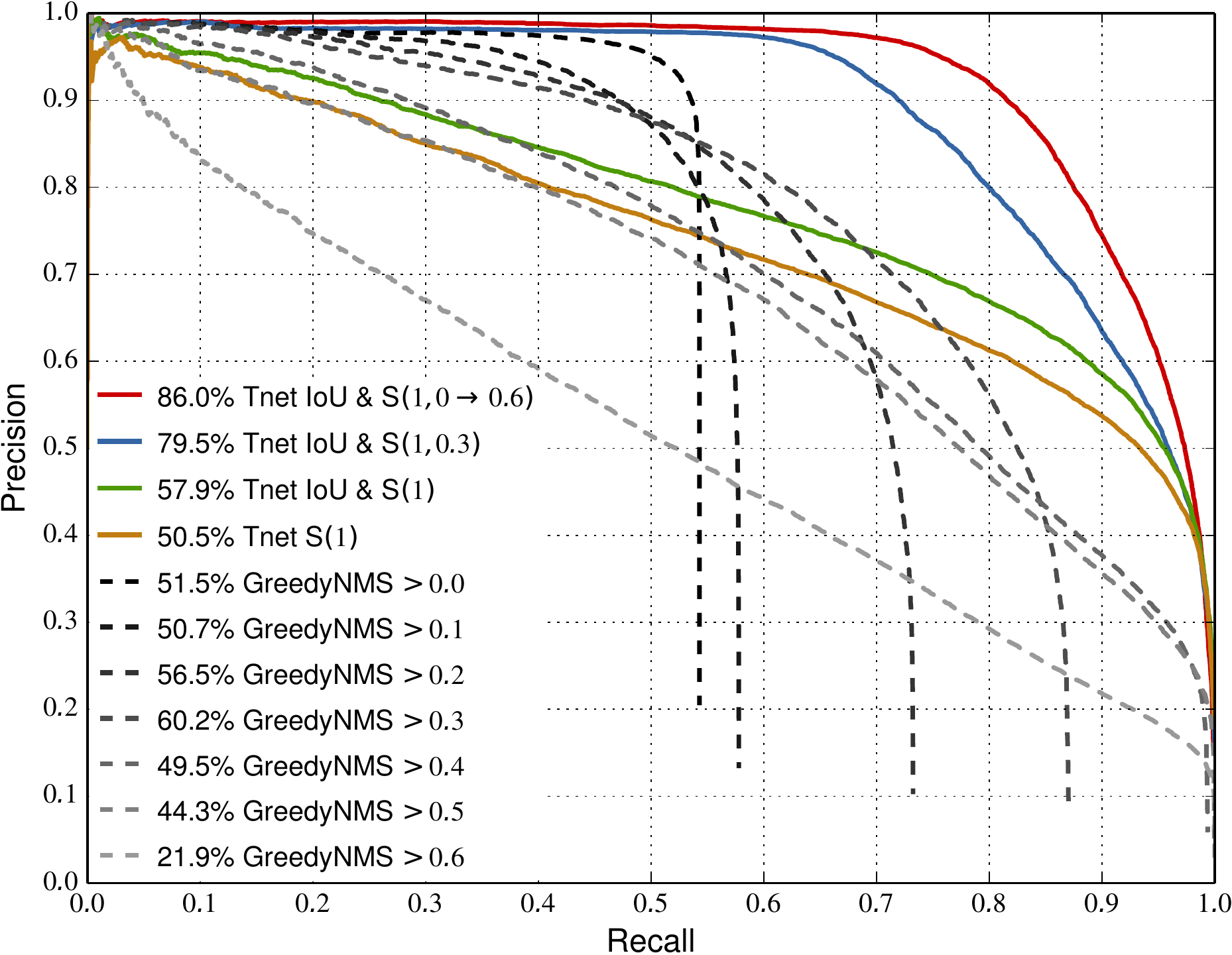}
\par\end{centering}

\caption{\label{fig:mnist-test-results}Detection results on controlled setup
(oMNIST test set).}
\end{figure}
\end{minipage}\hspace*{\fill}%
\begin{minipage}[t]{0.35\textwidth}%
\begin{table}[H]
\caption{\label{tab:mnist-results}Results from controlled setup experiments.
$\mbox{S}\left(\cdot\right)$ indicates the different input score
maps.}

\vspace{0.5em}

\begin{centering}
\begin{tabular}{lc}
Method & AR\tabularnewline
\hline 
\hline 
\vspace{-0.7em}
 & \tabularnewline
Upper bound & \tabularnewline
\quad{}perfect masks & 90.0\%\tabularnewline
\hline 
\vspace{-0.7em}
 & \tabularnewline
GreedyNMS  & \tabularnewline
\quad{}bboxes $\mbox{IoU}>0.3$ & 60.2\%\tabularnewline
\vspace{-0.7em}
 & \tabularnewline
Tnet & \tabularnewline
\quad{}$\mbox{IoU}\,\&\,\mbox{S}\negthinspace\left(1,\,0\rightarrow0.6\right)$ & \emph{86.0\%}\tabularnewline
\quad{}$\mbox{IoU}\,\&\,\mbox{S}\negthinspace\left(1,\,0.3\right)$ & 79.5\%\tabularnewline
\quad{}$\mbox{IoU}\,\&\,\mbox{S}\negthinspace\left(1\right)$ & 57.9\%\tabularnewline
\quad{}$\mbox{S}\negthinspace\left(1\right)$ & 50.5\%\tabularnewline
\end{tabular}
\par\end{centering}

\end{table}
\end{minipage}\hspace*{\fill}

\end{figure*}

Results are summarised in table \ref{tab:mnist-results} and figure
\ref{fig:mnist-test-results}. We summarise the curves via AR; the
average recall on the precision range $[0.5,\,1.0]$. The evaluation
is done using the standard Pascal VOC protocol, with $\mbox{IoU}>0.5$
\citep{Everingham2015Ijcv}.

\paragraph{GreedyNMS}

As can be easily seen in figure \ref{fig:mnist-test-results} varying
the $\mbox{IoU}$ thresholds for GreedyNMS trades off precision and
recall as discussed in section \ref{sec:Introduction}. The best AR
that can be obtained with GreedyNMS is $60.2\%$ for $\mbox{IoU}>0.3$.
Example score maps for this method can be found in figure \ref{fig:omnist-nms-tnet}.

\paragraph{Upper bound}

As an upper bound for any method relying on score map information
we can calculate the overlap between neighbouring hypotheses based
on perfect segmentation masks available in this toy scenario. In that
case even a simple strategy such as GreedyNMS can be used and based
on our idealized but noisy detection score maps this results in $90.0\%\ \mbox{AR}$.
In section \ref{sec:pets-experiments} we report experiments using
segmentation masks estimated from the image content that result in
inferior performance however.

\paragraph{Base Tnet}

Using the same information as GreedyNMS with bounding boxes, our base
Tnet reaches better performance for the entire recall range (see figure
\ref{fig:mnist-test-results} and table \ref{tab:mnist-results},
$\mbox{S}\negthinspace\left(1,\,0.3\right)$ indicates the score maps
from GreedyNMS with $\mbox{IoU}>0.3$ and $\geq1$, i.e. the raw score
map). In this configuration Tnet obtains $79.5\%\ \mbox{AR}$ clearly
superior to GreedyNMS. This shows that, at least in a controlled setup,
a convnet can indeed exploit the available information to overcome
the limitations of the popular GreedyNMS method.

Instead of picking a specific $\mbox{IoU}$ threshold to feed Tnet,
we consider $\mbox{IoU}\,\&\,\mbox{S}\negthinspace\left(1,\,0\rightarrow0.6\right)$,
which includes $\mbox{S}\negthinspace\left(1,\,0.6,\,0.4,\,0.3,\,0.2,\,0.0\right)$.
As seen in figure \ref{fig:mnist-test-results}, not selecting a specific
threshold results in the best performance of $86.0\%\ \mbox{AR}$.
As soon as some ranking signal is provided (via GreedyNMS results),
our Tnet is able to learn how to exploit best the information available.
 Qualitative results are presented in figure \ref{fig:omnist-nms-tnet}.

Table \ref{tab:mnist-results} reports results for a few degraded
cases. If we remove GreedyNMS $\mbox{S}\negthinspace\left(0.3\right)$
and only provide the raw score map ($\mbox{IoU}\,\&\,\mbox{S}\negthinspace\left(1,\,0.3\right)$)
performance decreases somewhat.

\paragraph{Auto-context}

Importantly we show that $\mbox{IoU}\,\&\,\mbox{S}\negthinspace\left(1\right)$
improves over $\mbox{S}\negthinspace\left(1\right)$ only. ($\mbox{S}\negthinspace\left(1\right)$
is the information exploited by auto-context methods, see \S\ref{sub:Related-work}).
This shows that the convnet is learning to do more than simple auto-context.
The detection improves not only by noticing patterns on the score
map, but also on how the detection boxes overlap.

\section{\label{sec:pets-experiments}Person detection experiments}

After the proof of concept for a controlled setup, we move to a realistic
pedestrian detection setup. We are particularly interested in datasets
that show diverse amounts of occlusion (and thus \NMS is non-trivial).
We decided for the PETS dataset, which exhibits diverse levels of
occlusion and provides a reasonable volume of training and test data.
Additionally we test the generalization of the trained model on the
ParkingLot dataset.\begin{wrapfigure}{O}{0.4\columnwidth}%
\vspace{-2.5em}

\begin{centering}
\includegraphics[width=0.4\textwidth]{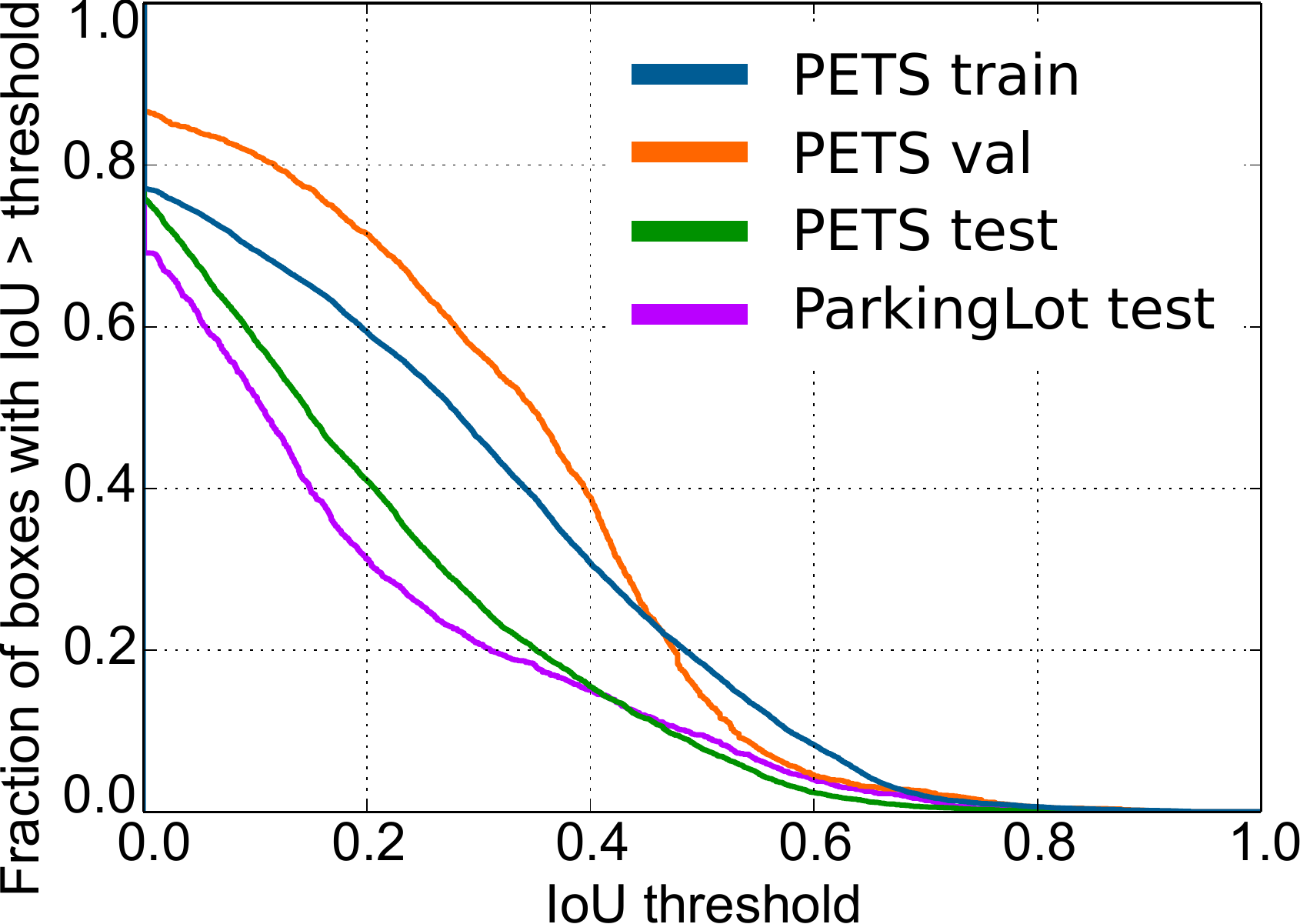}
\par\end{centering}

\caption{\label{fig:pets-gt-iou-distribution}Distribution of overlap between
ground truth PETS and ParkingLot annotations (measured in $\mbox{IoU}$).
Most persons on the datasets have some occlusion, and about $\sim\negmedspace20\%$
have significant occlusion ($\mbox{IoU}>0.4$).}

\vspace{-2em}
\end{wrapfigure}%

\paragraph{PETS}

We use 8 of the PETS sequences \citep{Ferryman2010AvssPetsDataset},
$\sim\negmedspace200$ frames each, that we split in 5 for training
(S1L1-1, S1L1-2, S1L2-1, S1L2-2, S2L1, and S3MF1), 1 for validation
(S2L3) and 1 for testing (S2L2). The different videos show diverse
densities of crowds. As shown in figure~\ref{fig:pets-gt-iou-distribution}
more than $40$/$50$/$25\%$ of the train/val/test data has an $\mbox{IoU}>0.3$
with another ground truth box. \\
Since detectors tend to have non-zero detection scores in most areas
of the image, the training volume is proportional to number of pixels
not the number of images. Thus we can adequately train our Tnet with
only a few hundred frames.\\
PETS has been previously used to study person detection \citep{Tang2013Iccv},
tracking \citep{Milan2014Pami}, and crowd density estimation \citep{Subburaman2012Avss}.
Standard pedestrian datasets such as Caltech \citep{Dollar2012Pami}
or KITTI \citet{Geiger2012Cvpr} average less than two pedestrian
per frame, making close-by detections a rare occurrence.\\
Due to its size and challenging occlusion statistics we consider PETS
a suitable dataset to explore \NMS. Figure \ref{fig:pets-qualitative-results}
shows example frames.

\paragraph{ParkingLot}

We use the first ParkingLot \citep{Shu2012Cvpr} sequence to evaluate
the generalization capabilities of the model. We use an improved set
annotations, provided every third frame (250 frames in total) and
rectify the mistakes from the original annotations. Compared to PETS
the sequence has similar overlap statistics than the PETS test set
(see figure~\ref{fig:pets-gt-iou-distribution}), but presents different
background and motion patterns. Figure \ref{fig:parkinglot-qualitative-results}
shows examples from the dataset.

\paragraph{Person detector}

In this work we take the detector as a given. For our experiments
we use the baseline DPM detector from \citep{Tang2013Iccv}. We are
not aware of a detector (convnet or not) providing better results
on PETS-like sequences (we considered some of the top detectors in
\citep{Dollar2012Pami}). Importantly, for our exploration the detector
quality is not important per-se. As discussed in section \ref{sub:oMNIST-dataset}
GreedyNMS suffers from intrinsic issues, even when providing an idealized
detector. In fact Tnet benefits from better detectors, since there
will be more signal in the score maps, it becomes easier to do \NMS.
We thus consider our DPM detector a fair detection input.\\
We use the DPM detections after bounding box regression, but before
any \NMS processing.

\paragraph{Person segments}

In section \ref{sub:PETS-results} we report results using segments
estimated from the image content. We use our re-implementation of
DeepMask \citep{Pinheiro2015ArxivDeepMask}, trained on the Coco
dataset \citep{Lin2014EccvCoco}. DeepMask is a network specifically
designed for objects segmentation which provides competitive performance.
Our re-implementation obtains results of comparable quality as the
original; example results on PETS are provided in the appendix section
\ref{sec:DeepMask-on-PETS}. We use DeepMask as a realistic example
of what can be expected from modern techniques for instance segmentation.

\subsection{\label{sub:PETS-results}PETS results}

\begin{figure*}[t]

\hspace*{\fill}%
\begin{minipage}[t]{0.4\textwidth}%
\begin{table}[H]
\caption{\label{tab:pets-test-results}Results on PETS validation set. Underlined
is our base Tnet.}

\vspace{0.5em}

\centering{}%
\begin{tabular}{lc}
Method & AR\tabularnewline
\hline 
\hline 
GreedyNMS  & \tabularnewline
\quad{}bboxes $\mbox{IoU}>0.3$ & 54.3\%\tabularnewline
\quad{}DeepMask segments & 52.0\% \tabularnewline
\vspace{-0.7em}
 & \tabularnewline
Tnet variants & \tabularnewline
\quad{}$\mbox{IoU}\,\&\,\mbox{S}\negthinspace\left(1,\,0\negthinspace\rightarrow\negthinspace0.6\right)$ & \emph{59.6\%}\tabularnewline
\quad{}$\mbox{IoU}\,\&\,\mbox{S}\negthinspace\left(1,\,0.3\right)$,
$r=0.3$\hspace*{-0.5em} & 58.9\%\tabularnewline
\quad{}$\mbox{IoU}\,\&\,\mbox{S}\negthinspace\left(1,\,0.3\right)$,
$w_{g}$ & 58.0\%\tabularnewline
\quad{}$\mbox{IoU}\,\&\,\mbox{S}\negthinspace\left(1,\,0.3\right)$ & \uline{57.9\%}\tabularnewline
\quad{}$\mbox{IoU}\,\&\,\mbox{S}\negthinspace\left(1\right)$ & 36.5\%\tabularnewline
\quad{}$\mbox{S}\negthinspace\left(1\right)$ & 33.9\%\tabularnewline
\end{tabular}
\end{table}
\end{minipage}\hspace*{\fill}%
\begin{minipage}[t]{0.6\textwidth}%
\begin{figure}[H]
\begin{centering}
\includegraphics[width=1\textwidth]{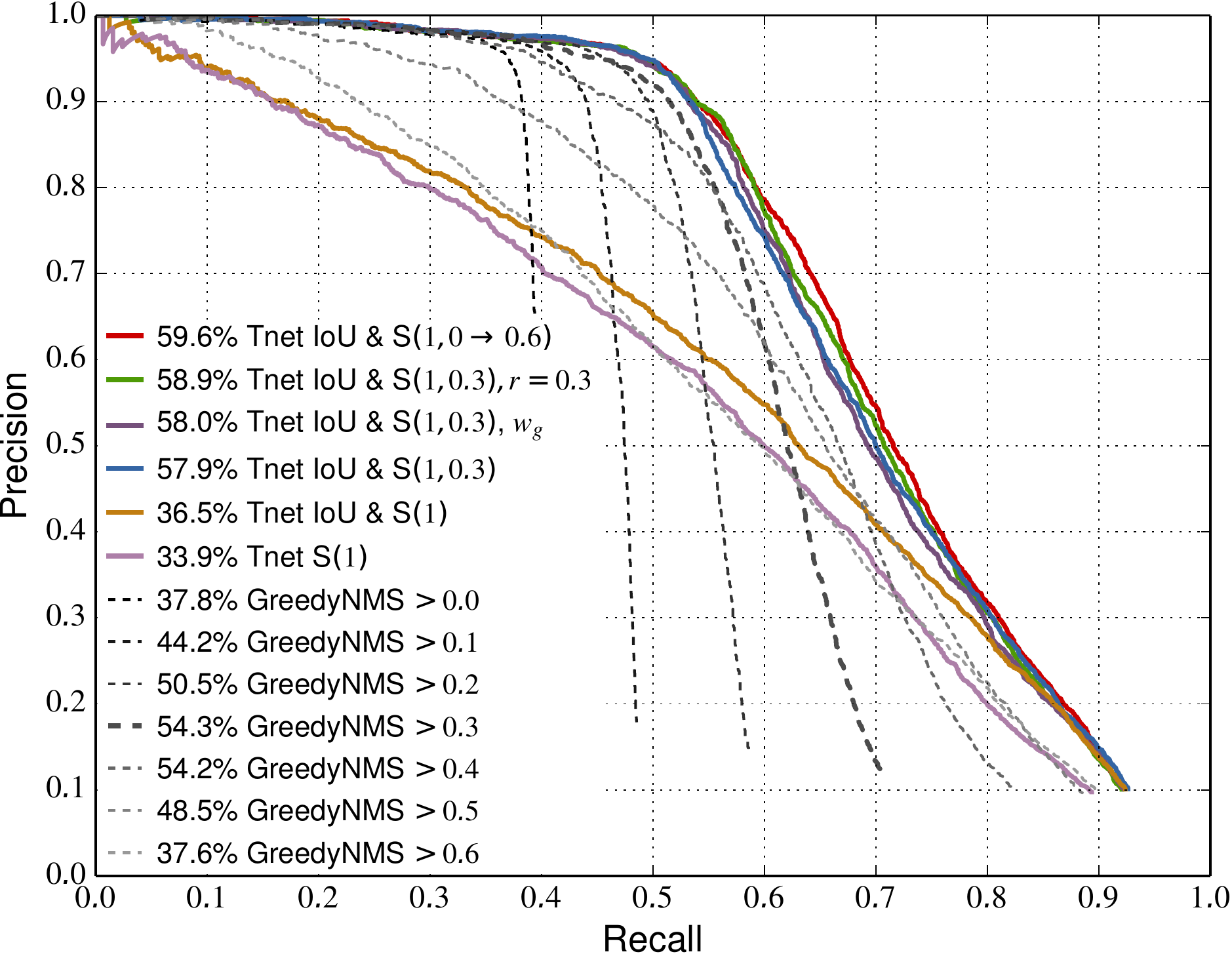}
\par\end{centering}

\caption{\label{fig:pets-val-results}Detection results on PETS validation
set. Global weighting is indicated by $w_{g}$, all other curves use
frame weighting.}
\end{figure}
\end{minipage}\hspace*{\fill}

\end{figure*}

\begin{figure}
\centering{}\hspace*{\fill}%
\begin{minipage}[t]{0.47\columnwidth}%
\begin{center}
\includegraphics[width=1\textwidth]{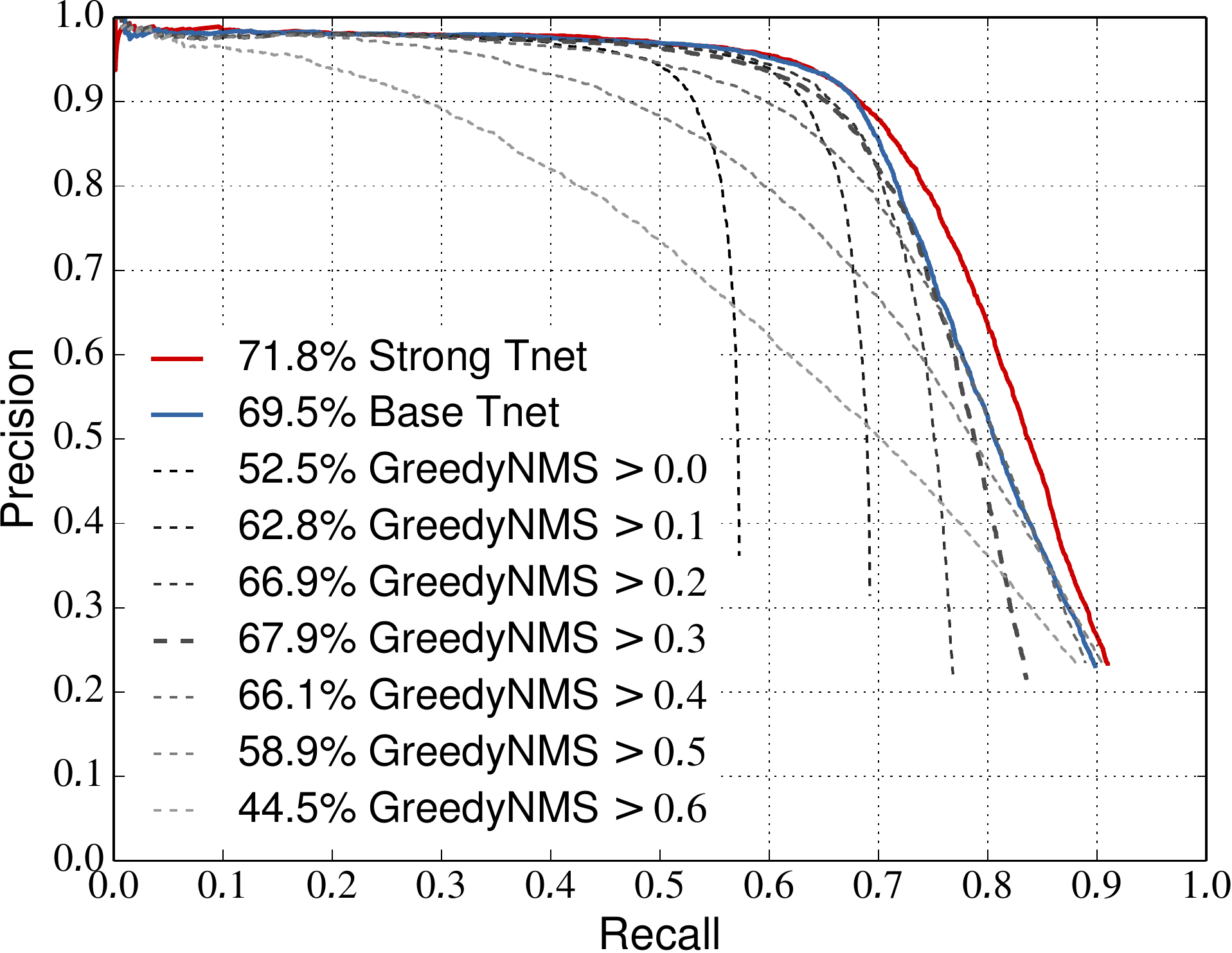}
\par\end{center}

\caption{\label{fig:pets-test-results}Detection results on PETS test set.
Our approach is better than any GreedyNMS threshold and better than
the upper envelope of all GreedyNMS curves.}
\end{minipage}\hspace*{\fill}%
\begin{minipage}[t]{0.47\columnwidth}%
\begin{center}
\includegraphics[width=1\textwidth]{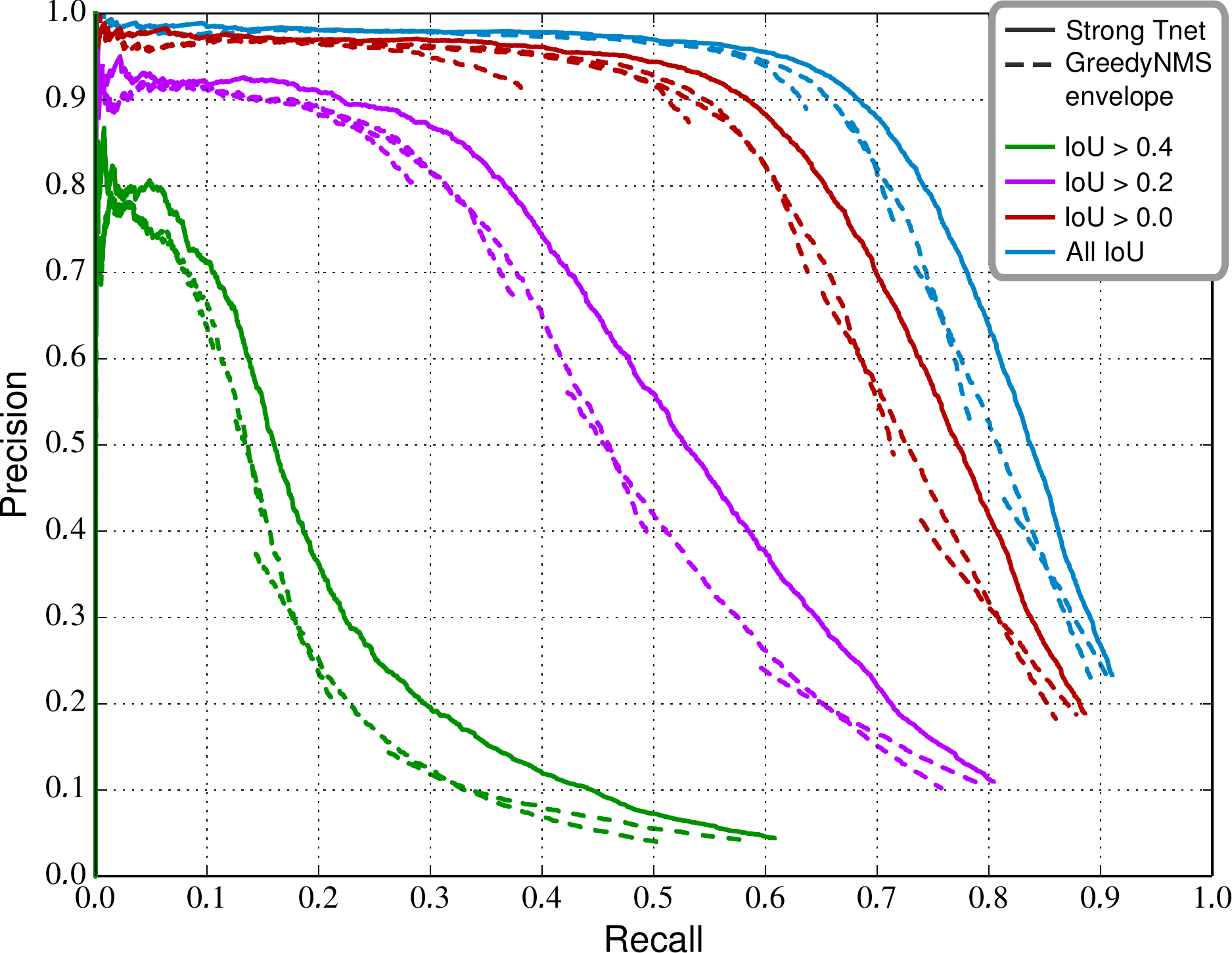}
\par\end{center}

\caption{\label{fig:tnet-on-test-subsets}GreedyNMS versus Strong Tnet when
evaluated over different subsets of PETS test data, based on level
of occlusion. In each subset our Tnet improves over the upper envelope
of all GreedyNMS threshold curves.}
\end{minipage}\hspace*{\fill}
\end{figure}

Our PETS results are presented in table \ref{tab:pets-test-results},
figure \ref{fig:pets-val-results} (validation set), figures \ref{fig:pets-test-results},
\ref{fig:pets-test-results} (test set). Qualitative results are shown
in figure \ref{fig:pets-qualitative-results}.

\paragraph{GreedyNMS on boxes}

Just like in the oMNIST case, the GreedyNMS curves in figure \ref{fig:pets-val-results}
have a recall versus precision trade-off. For PETS we pick $\mbox{IoU}>0.3$
as a reference threshold which provides a reasonable balance.

\paragraph{GreedyNMS on segments}

As discussed in section \ref{sec:mnist-experiments}, GreedyNMS should
behave best when the detection overlap is based on the visible area
of the object. We compute DeepMask segmentations on the DPM detection,
feed these in GreedyNMS, and select the best $\mbox{IoU}$ threshold
for the validation set. As seen in table \ref{tab:pets-test-results}
results are slightly under-performing the bounding boxes cases. Albeit
many segments are rather accurate (see appendix section \ref{sec:DeepMask-on-PETS}),
segments tend to be accurate on the easy cases for the detector, and
drop in performance when heavier occlusion is present. Albeit in
theory having good segmentations should make GreedyNMS quite better,
in practice they do not. At the end, the segments hurt more than they
help for GreedyNMS.

\paragraph{Auto-context}

The entry $\mbox{S}\left(1\right)$ in table \ref{tab:pets-test-results}
shows the case where only the raw detection score map is feed to Tnet
(same nomenclature as section \ref{sub:mnist-results}). Since performance
is lower than other variants (e.g. $\mbox{IoU}\,\&\,\mbox{S}\left(1\right)$),
this shows that our approach is exploiting available information better
than just doing auto-context over DPM detections.

\paragraph{Tnet}

Both in validation and test set (figures \ref{fig:pets-val-results}
and \ref{fig:pets-test-results}) our trained network with $\mbox{IoU}\,\&\,\mbox{S}\left(1,\,0.3\right)$
input provides a clear improvement over vanilla GreedyNMS. Just like
in the oMNIST case, the network is able to leverage patterns in the
detector output to do better \NMS than the de-facto standard GreedyNMS
method.\\
Table \ref{tab:pets-test-results} report the results for a few additional
variants. $\mbox{IoU}\,\&\,\mbox{S}\negthinspace\left(1,\,0\rightarrow0.6\right)$
shows that it is not necessary to select a specific $\mbox{IoU}$
threshold for the input score map layer. Given an assortment ($\mbox{S}\negthinspace\left(1,\,0.6,\,0.4,\,0.3,\,0.2,\,0.0\right)$)
the network will learn to leverage the information available.\\
Using the relaxed loss described in \ref{sub:Training} helps further
improve the results. Amongst the parameters tried on the validation
set, $r=0.3$ provides the largest improvement. Lower $r$ values
decrease performance, while higher $r$ values converge towards the
default $r=1$ performance (base Tnet).\\
Weighting classes equally on the entire dataset ($w_{g}$ strategy)
gives a mild improvement from $57.9\%$ to $58.0$\% AR compared to
the default per frame weighting $w_{f}$. 

\paragraph{Strong Tnet}

We combine the best ingredients identified on the validation set into
one strong model. We use $\mbox{IoU}\,\&\,\mbox{S\ensuremath{\left(1,\,0\negthinspace\rightarrow\negthinspace0.6\right)}}$,
relaxed loss with $r=0.3$, and global weighting $w_{g}$. Figure
\ref{fig:pets-test-results} shows that we further improve over the
base Tnet from $59.5\%$ to $71.8\%\ \mbox{AR}$ on the PETS test
set. The gap between base Tnet and GreedyNMS is smaller on the test
set than on validation, because test set has lighter occlusions (see
figure \ref{fig:pets-gt-iou-distribution}). Still our strong Tnet
provides a consistent improvement over GreedyNMS.\\
Figure \ref{fig:tnet-on-test-subsets} provides a more detailed view
of the results from figure \ref{fig:pets-test-results}. It compares
our strong Tnet result versus the upper envelope of GreedyNMS over
all thresholds ($[0,\,1]$), when evaluated over different subsets
of the test set. Each subset corresponds to ground truth bounding
boxes with other boxes overlapping more than a given $\mbox{IoU}$
level (see figure \ref{fig:pets-gt-iou-distribution}). For all ranges,
our strong Tnet improves over GreedyNMS. This shows that our network
does not fit to a particular range of occlusions, but learns to handle
all of them with comparable effectiveness.\\
At test time Tnet is reasonably fast, taking $\sim\negmedspace200\ \mbox{milliseconds}$
per frame (all included).

\subsection{ParkingLot results}

\begin{wrapfigure}{O}{0.41\columnwidth}%
\vspace{-4em}

\begin{centering}
\includegraphics[bb=0bp 0bp 568bp 432bp,width=0.4\textwidth]{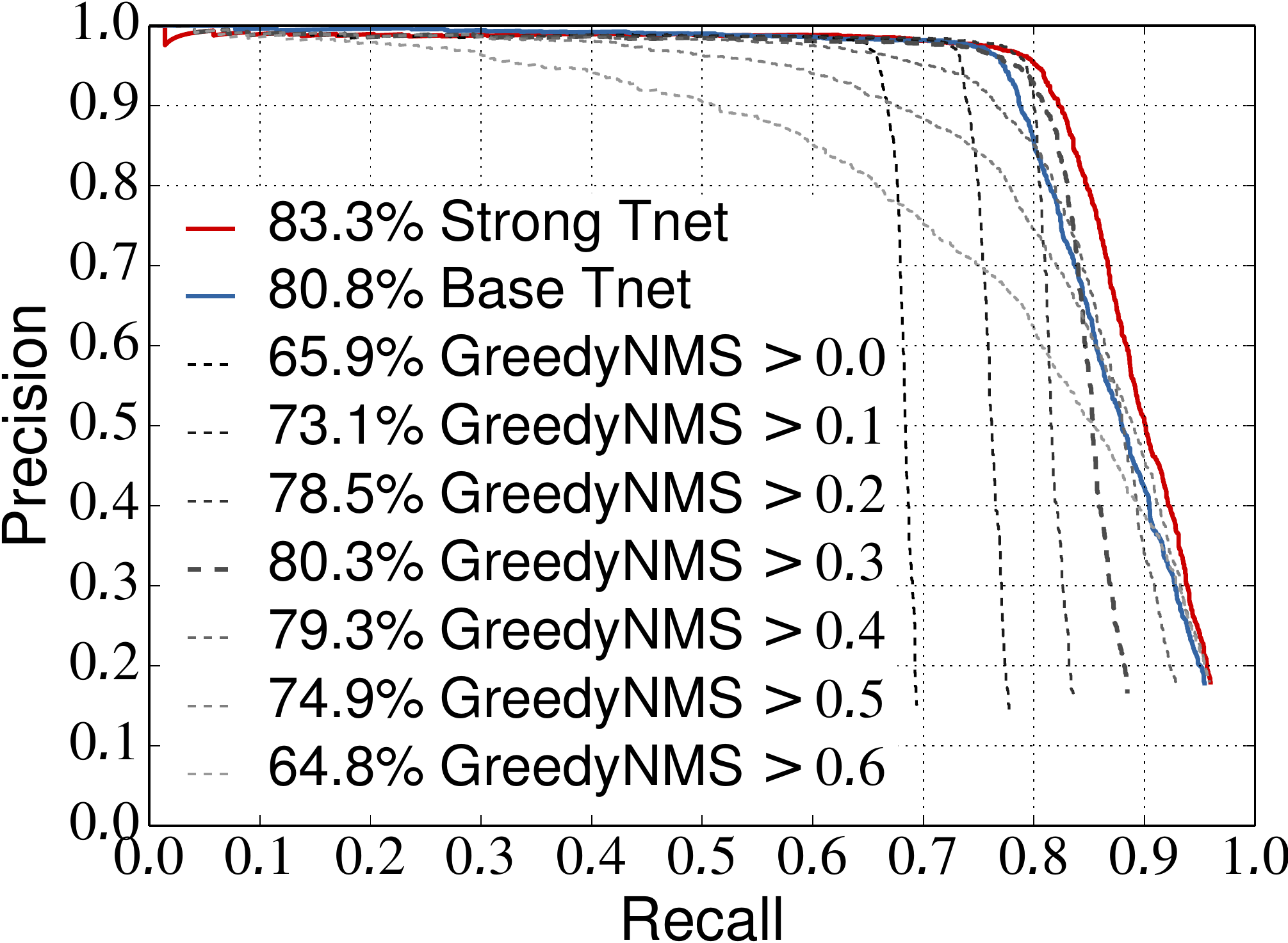}
\par\end{centering}

\caption{\label{fig:parking-lot-results}Detection results on the ParkingLot
dataset. Tnet is better than any GreedyNMS threshold, even though
it has been trained using PETS data only.}

\vspace{-4em}
\end{wrapfigure}%

To verify that our Tnet can generalize beyond PETS, we run the same
DPM detector as on the PETS experiment over the ParkingLot sequence
and do NMS using the networks trained on PETS training set only. Results
in figure~\ref{fig:parking-lot-results} show that Tnet improves
from $80.3\%$ to $83.3\%$ AR over the best GreedyNMS threshold of
$\text{{IoU}}>0.3$. Even though Tnet was not trained on this sequence
we see a similar result as on the PETS dataset. Not only does our
Strong Tnet improve over the best GreedyNMS result, but it improves
over the upper envelope of all GreedyNMS thresholds. See qualitative
results in figure~\ref{fig:parkinglot-qualitative-results}.

\section{\label{sec:Conclusion}Conclusion}

We have discussed in detail the limitations of GreedyNMS and presented
experimental examples showing its recall versus precision trade-off.
For the sake of speed and simplicity GreedyNMS disregards most of
the information available in the detector response. Our proposed Tyrolean
network (Tnet) mines the patterns in the score map values and bounding
box arrangements to surpass the performance of GreedyNMS. On the person
detection task, our final results show that our approach provides,
compared to any GreedyNMS threshold, both high recall and improved
precision. These results confirm that Tnet overcomes the intrinsic
limitations of GreedyNMS, while keeping practical test time speeds.

Albeit the proposed architecture results in good results for the scenario
covered, there is certainly room for further probing the parameter
space of convnets for \NMS and exploring other applications domains
(e.g. \NMS for boundaries estimation, or other detection datasets).
Explicitly handling detection scales, considering multi-class problems,
or back-propagating into the detector itself are future directions
of interest.

Current detection pipelines start with a convnet and end with a hard-coded
\NMS procedure. For detection problems where occlusions are present,
we reckon that significant improvements can be obtained by ending
with a Tnet.

\begin{figure}[h]
\begin{centering}
\hspace*{\fill}\subfloat[GreedyNMS]{\begin{centering}
\begin{tabular}{c}
\includegraphics[width=0.45\textwidth]{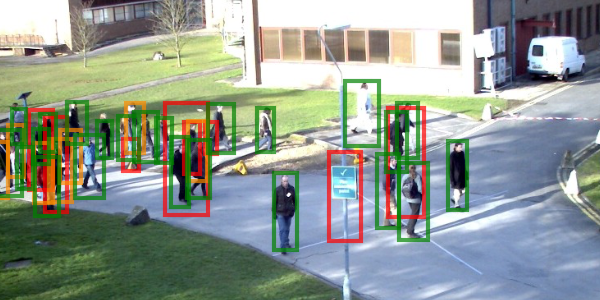}\tabularnewline
\includegraphics[width=0.45\textwidth]{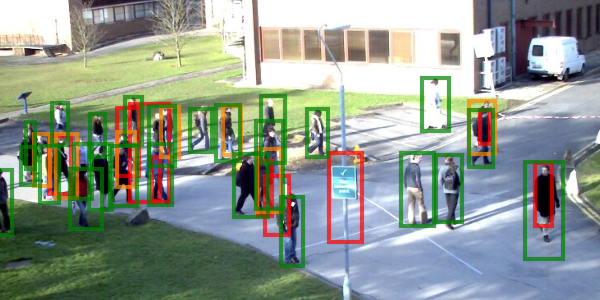}\tabularnewline
\includegraphics[width=0.45\textwidth]{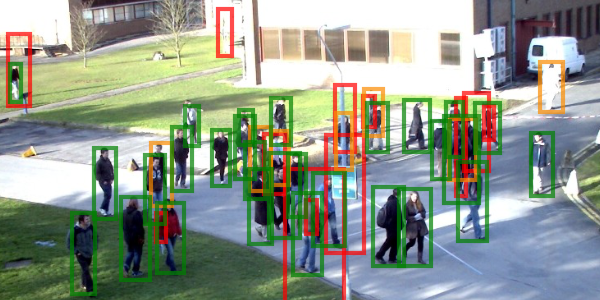}\tabularnewline
\end{tabular}
\par\end{centering}

}\hspace*{\fill}\subfloat[Strong Tnet]{%
\begin{tabular}{c}
\includegraphics[width=0.45\textwidth]{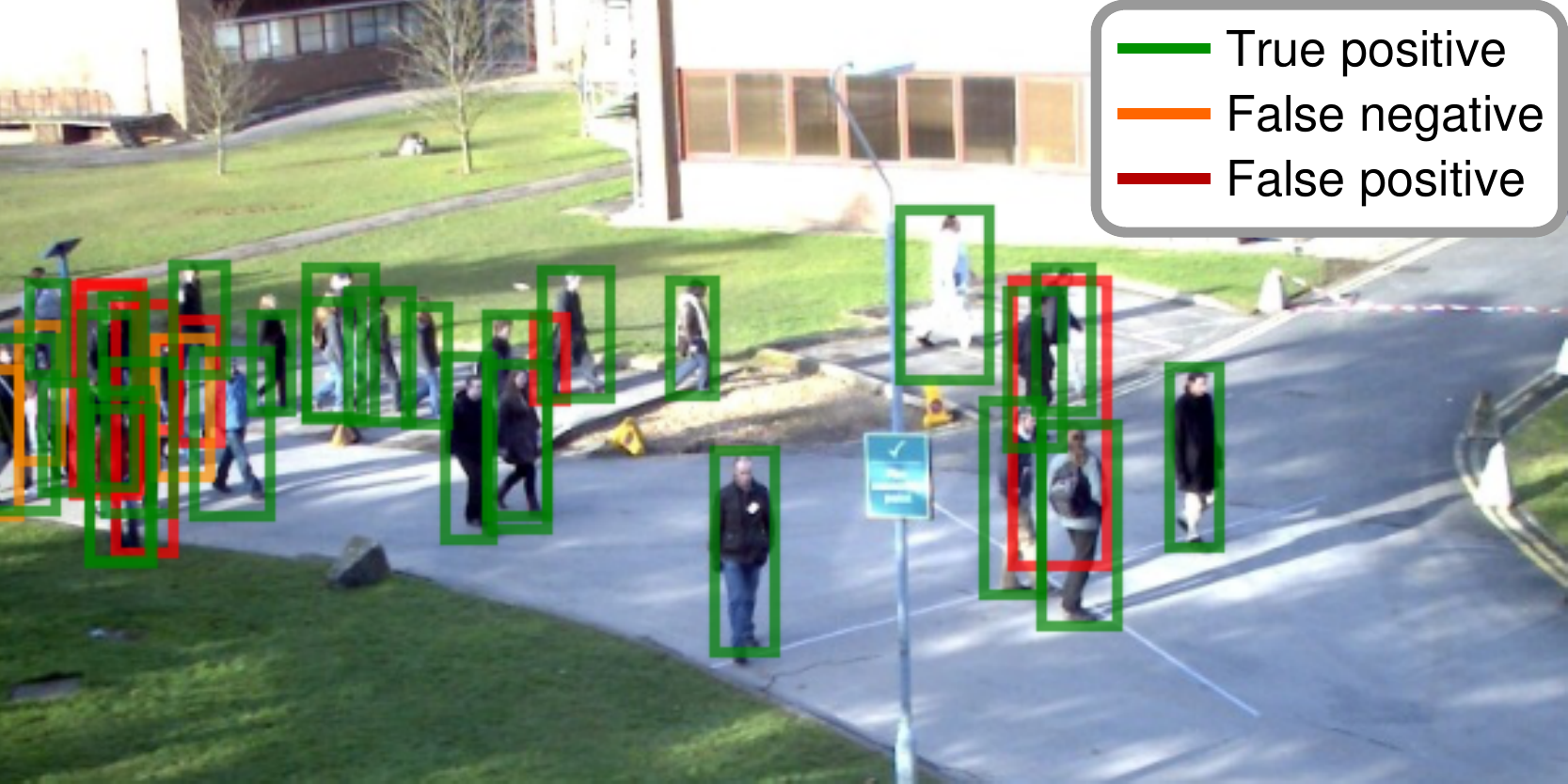}\tabularnewline
\includegraphics[width=0.45\textwidth]{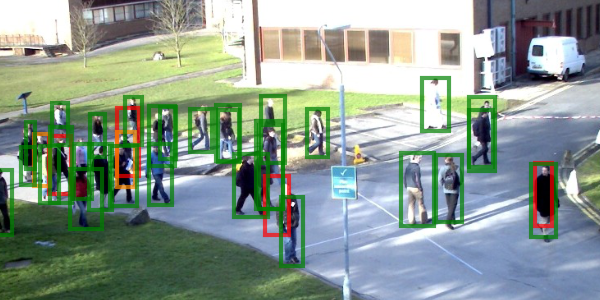}\tabularnewline
\includegraphics[width=0.45\textwidth]{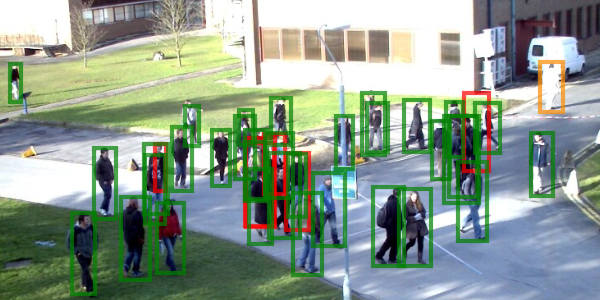}\tabularnewline
\end{tabular}}\hspace*{\fill}
\par\end{centering}

\begin{centering}

\par\end{centering}

\caption{\label{fig:pets-qualitative-results}Qualitative detection results
of GreedyNMS $>0.3$ and Strong Tnet over DPM detections on PETS test
set (both operating at 75\% recall). Tnet is able to suppress false
positives as well as recover recall that is lost with GreedyNMS. This
is the case for both crowded and non-crowded areas.}
\end{figure}
\begin{figure}[h]
\begin{centering}
\hspace*{\fill}\subfloat[GreedyNMS]{\begin{centering}
\begin{tabular}{c}
\includegraphics[width=0.45\textwidth]{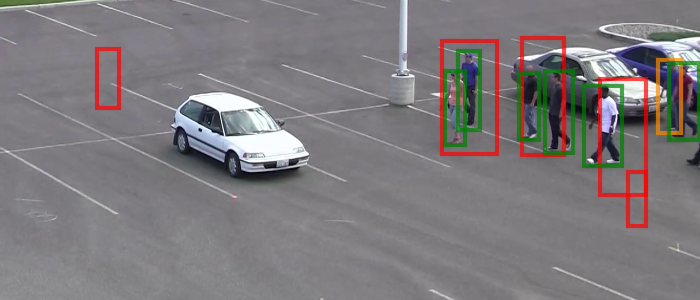}\tabularnewline
\includegraphics[width=0.45\textwidth]{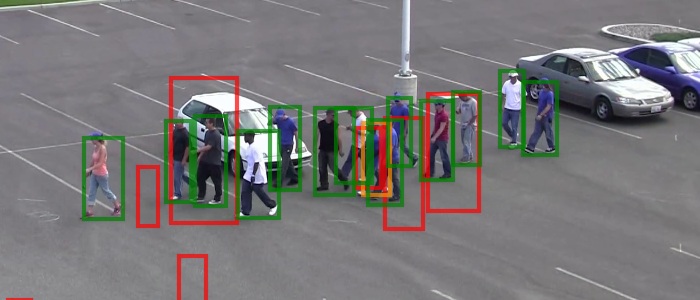}\tabularnewline
\includegraphics[width=0.45\textwidth]{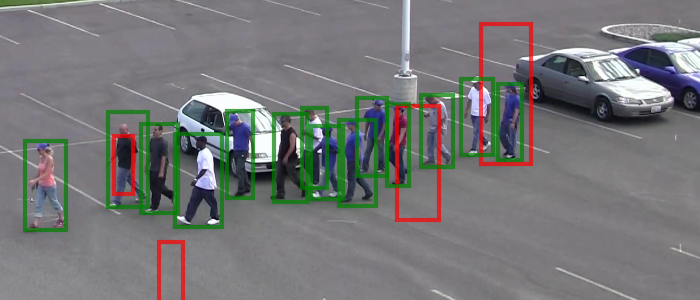}\tabularnewline
\end{tabular}
\par\end{centering}

}\hspace*{\fill}\subfloat[Strong Tnet]{%
\begin{tabular}{c}
\includegraphics[width=0.45\textwidth]{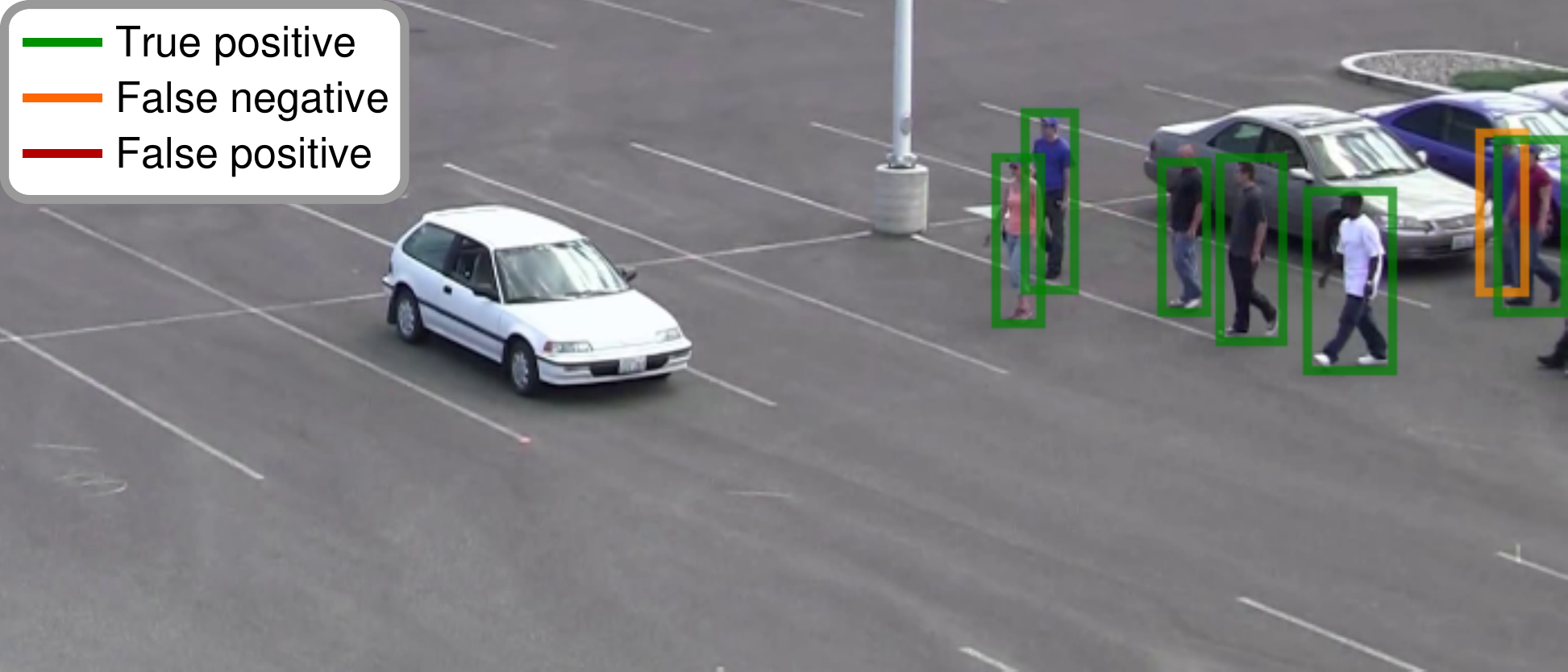}\tabularnewline
\includegraphics[width=0.45\textwidth]{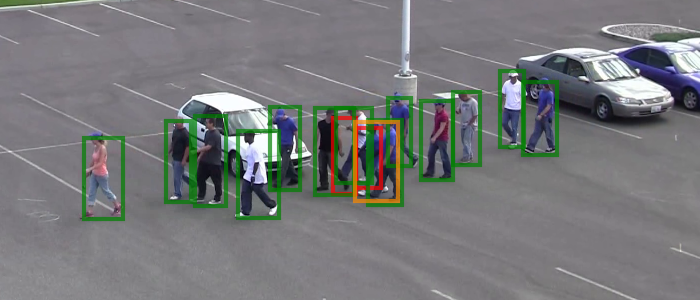}\tabularnewline
\includegraphics[width=0.45\textwidth]{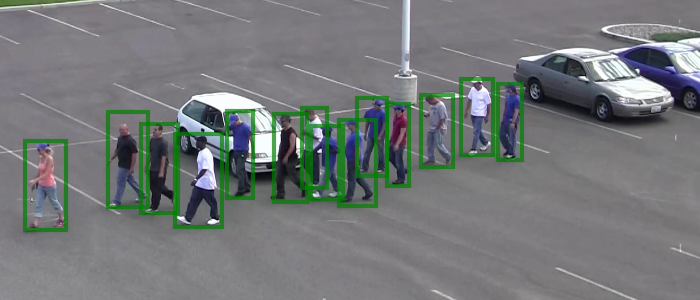}\tabularnewline
\end{tabular}}\hspace*{\fill}
\par\end{centering}

\begin{centering}

\par\end{centering}

\caption{\label{fig:parkinglot-qualitative-results}Qualitative detection results
of GreedyNMS $>0.3$ and Strong Tnet over DPM detections on ParkingLot
dataset (both operating at 85\% recall). Tnet was trained using PETS
only. }
\end{figure}

\subsubsection*{Acknowledgements}

We thank Siyu Tang and Anton Milan for providing the pre-trained DPM
model and additional PETS ground-truth annotations, respectively.

\bibliographystyle{iclr2016_conference}
\bibliography{2016_iclr_nms_network}

\clearpage{}

\appendix

\part*{Supplementary material}

\section{\label{sec:DeepMask-on-PETS}DeepMask results on PETS}

\begin{figure}[h]
\begin{centering}
\hspace*{\fill}\includegraphics[width=0.3\textwidth]{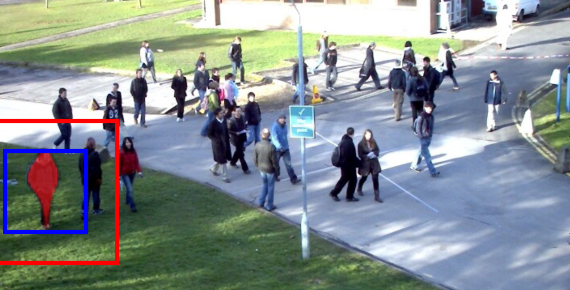}\hspace*{\fill}\includegraphics[width=0.3\textwidth]{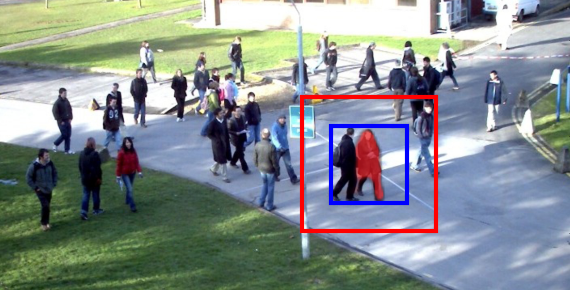}\hspace*{\fill}\includegraphics[width=0.3\textwidth]{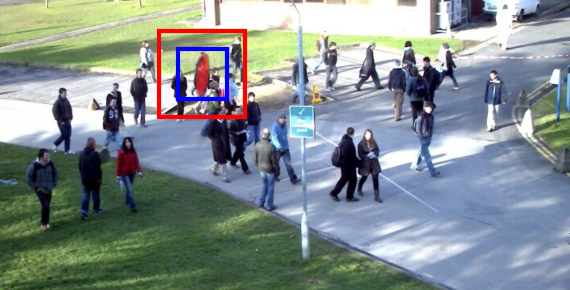}\hspace*{\fill}
\par\end{centering}

\begin{centering}
\hspace*{\fill}\includegraphics[width=0.3\textwidth]{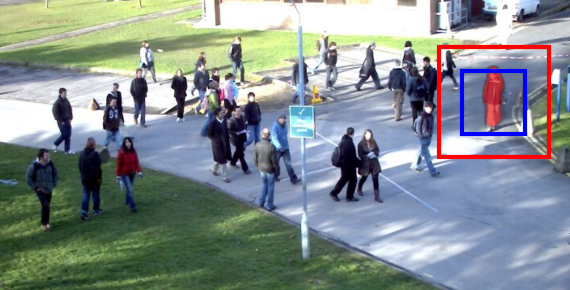}\hspace*{\fill}\includegraphics[width=0.3\textwidth]{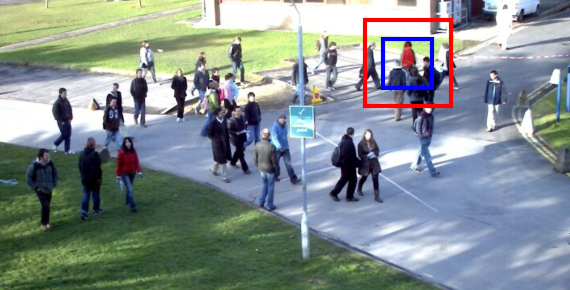}\hspace*{\fill}\includegraphics[width=0.3\textwidth]{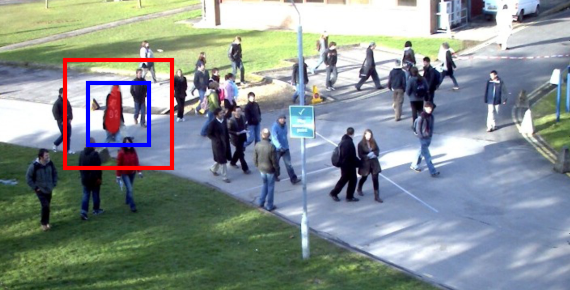}\hspace*{\fill}
\par\end{centering}

\begin{centering}
\hspace*{\fill}\includegraphics[width=0.3\textwidth]{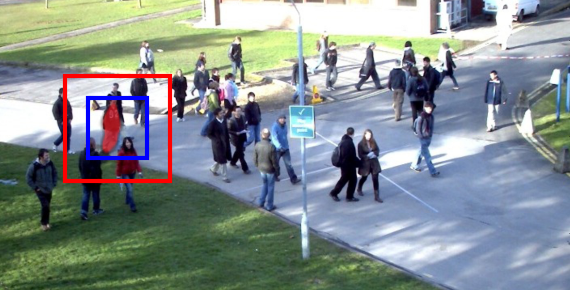}\hspace*{\fill}\includegraphics[width=0.3\textwidth]{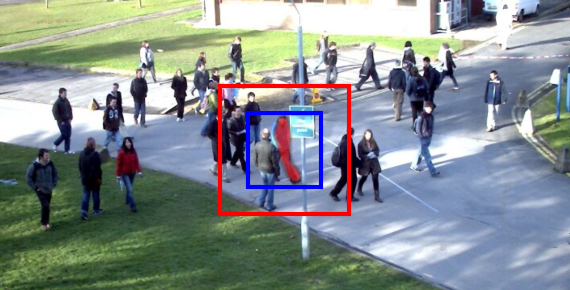}\hspace*{\fill}\includegraphics[width=0.3\textwidth]{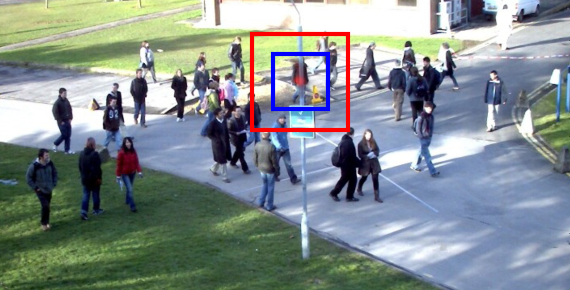}\hspace*{\fill}
\par\end{centering}

\caption{\label{fig:deepmask-pets}Example DeepMask segmentation masks on PETS
images. Pixels inside the red area are used to predict a foreground
segment inside the blue area. In these examples, boxes are centred
on ground truth annotations.}
\end{figure}

To obtain segmentation masks on PETS, we train our reimplementation
of DeepMask \citep{Pinheiro2015ArxivDeepMask} for all classes on
the COCO training set. Our implementation is based on the Fast-RCNN
network \citet{Girshick2015IccvFastRCNN}. To generate instance segmentations
on PETS, we upscale the image by a factor of $2\times$ and predict
segments on detections. Figure \ref{fig:deepmask-pets} shows mask
predictions for annotations on the PETS test set. It works well in
low occlusion cases (left and middle column), however, under heavy
occlusion it makes mistakes by collapsing the segment or merging the
occluding and the occluded person (see right-most column).

\end{document}